\documentclass[11pt]{article}

\usepackage[final]{acl}

\usepackage{times}
\usepackage{latexsym}

\usepackage[T1]{fontenc}

\usepackage[utf8]{inputenc}

\usepackage{microtype}

\usepackage{inconsolata}

\usepackage{graphicx}

\usepackage{kotex}
\usepackage{amsmath}
\usepackage{booktabs}
\usepackage[capitalize]{cleveref}
\crefname{section}{Sec.}{Secs.}
\Crefname{section}{Section}{Sections}
\Crefname{table}{Table}{Tables}
\crefname{table}{Tab.}{Tabs.}
\usepackage{amssymb}
\usepackage{comment}
\usepackage{subcaption}
\usepackage{multirow}
\usepackage{multicol}
\usepackage{wrapfig}
\usepackage{pgfplots}
\usepackage[table]{xcolor}
\pgfplotsset{compat=1.18}

\title{LoopCD: Loop-wise Contrastive Decoding for Improving Reasoning in Looped Language Models}

\author{
  Byeongho Yu$^{1}$$^{*}$ \quad 
  Junhyuk So$^{1}$$^{*}$ \quad 
  Eunhyeok Park$^{2}$\\[0.3em]
  $^{1}$Department of Computer Science and Engineering \\
  $^{2}$Graduate School of Artificial Intelligence \\
  Pohang University of Science and Technology (POSTECH) \\[0.3em]
  \texttt{\{bhyu418, junhyukso, eh.park\}@postech.ac.kr}
}

\begin{document}
\maketitle
\renewcommand{\thefootnote}{\fnsymbol{footnote}}
\footnotetext[1]{ These authors contributed equally.}
\renewcommand{\thefootnote}{\arabic{footnote}}
\begin{abstract}

Looped Language Models (LoopLMs) perform “latent reasoning” by recursively refining internal latent representations with shared weights, offering a more effective alternative to explicit verbal reasoning. 
Despite their effectiveness, we find that LoopLMs remain prone to \textit{loop instability}: unstable refinement across iterations can produce localized uncertain “hard” tokens associated with reasoning errors.
To address this, we propose \textbf{LoopCD}, loop-wise contrastive decoding that enhances the reasoning performance of LoopLMs by intervening on these tokens at inference time. 
Specifically, we exploit the internal dynamics of LoopLMs and contrast the logits from earlier iterations with logits from the last refined iteration to form the final sampling distribution. 
We find that this strategy is highly efficient, introducing only negligible inference overhead and requiring no additional training, while effectively improving reasoning performance by naturally refining reasoning-critical hard tokens.
Extensive experiments show that our method improves the performance of recent representative LoopLMs across various reasoning tasks.

\end{abstract}

\section{Introduction}
Recent advances in large language models (LLMs) have shown that scaling both model size \cite{singh2025openai, liu2024deepseek} and inference-time computation through \textit{verbal thinking}, such as chain-of-thought \cite{CoT}, can substantially enhance reasoning capabilities. However, this paradigm can be inherently inefficient: models require massive parameter capacity to support different stages of inference \cite{stageofinference}, and  verbalized thinking in discrete tokens imposes a linguistic bottleneck that leads to information collapse of rich continuous space \cite{hao2024training}.

\begin{figure}[t]
    \centering    \includegraphics[width=0.9\linewidth]{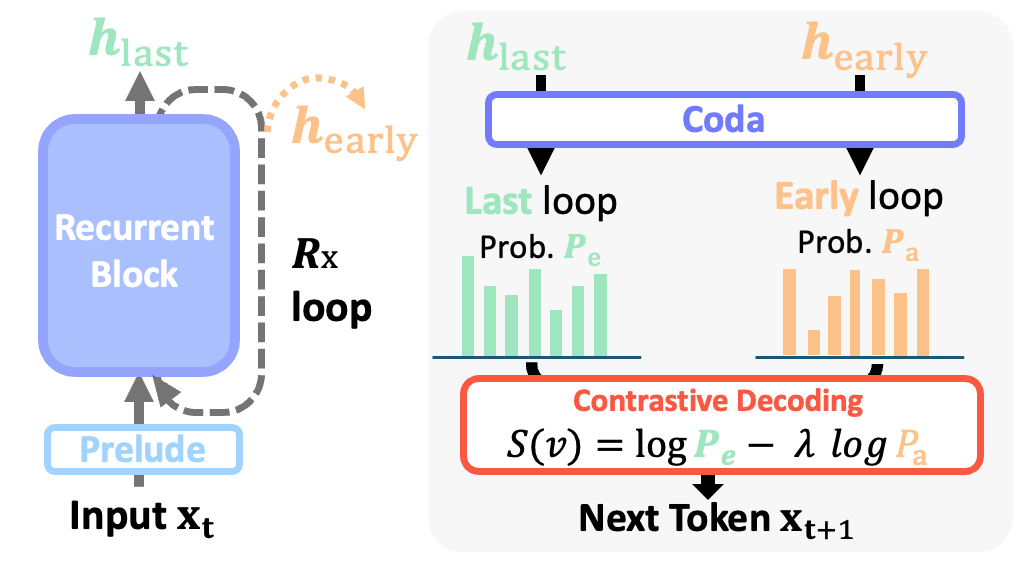}\vspace{-2mm}
    \caption{
    Overview of LoopCD at each decoding step.
    }
    \vspace{-5mm}
    \label{fig:fig1}
\end{figure}

As an emerging alternative, Looped Language Models (LoopLMs) \cite{huginn, ouro} have been explored for \textit{latent reasoning}, where models perform recursive computation directly in hidden space with shared weights, more naturally supporting different stages of inference and allowing models to reason in a richer continuous semantic space \cite{CTM}. Recent studies have shown that this design yields strong parameter efficiency \cite{ouro} and strong performance across diverse reasoning tasks.

Despite their effectiveness, we observe \textit{loop instability} in LoopLMs:
for a small subset of tokens, additional latent iterations do not monotonically improve predictions, but can induce unstable dynamics such as premature confidence, failed revision, or erroneous refinements.
Crucially, we find that this instability is often concentrated around reasoning-critical \textit{hard} tokens, where localized errors can accumulate along the reasoning trajectory and eventually propagate to the final answer.

\begin{figure*}[t]
    \centering

    \begin{subfigure}[t]{0.32\linewidth}
        \centering
        \includegraphics[width=\linewidth]{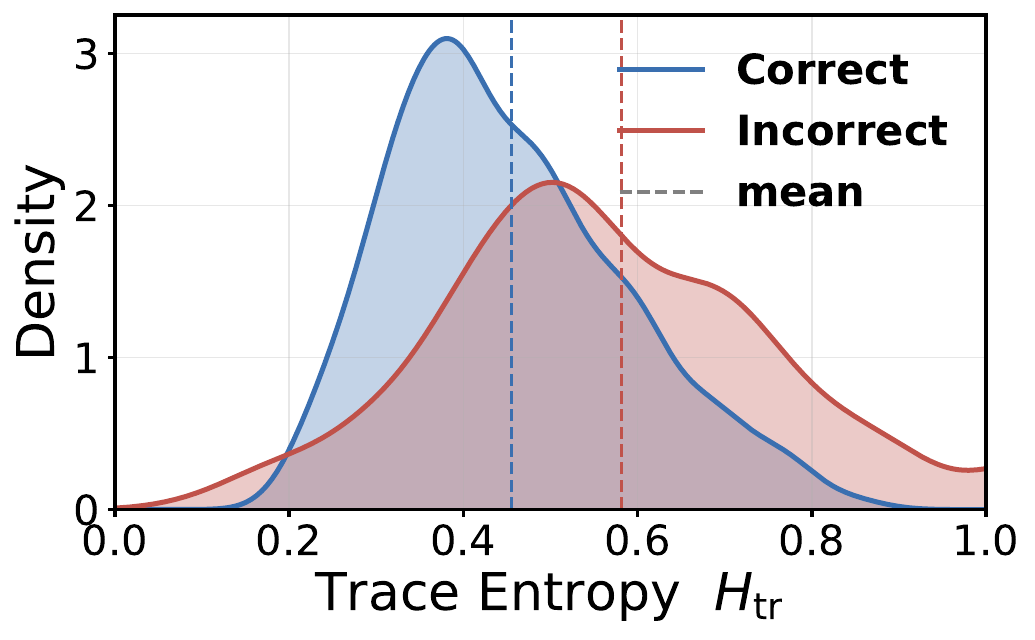}
        \caption{}
        \label{fig:main_a}
    \end{subfigure}
    \hfill
    \begin{subfigure}[t]{0.32\linewidth}
        \centering
        \includegraphics[width=\linewidth]{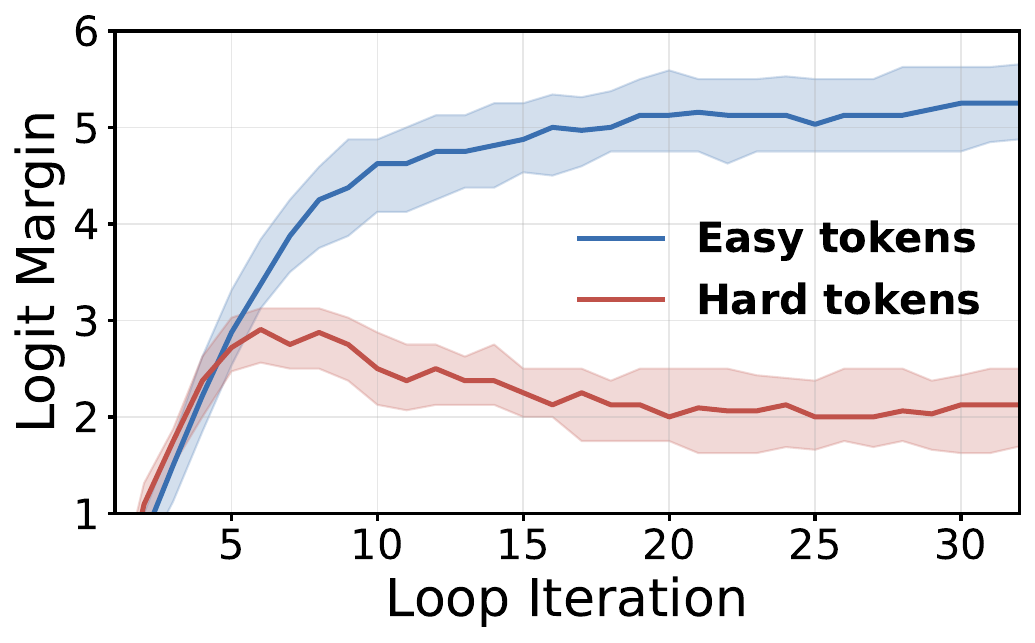}
        \caption{}
        \label{fig:main_b}
    \end{subfigure}
    \hfill
    \begin{subfigure}[t]{0.32\linewidth}
        \centering
        \includegraphics[width=\linewidth]{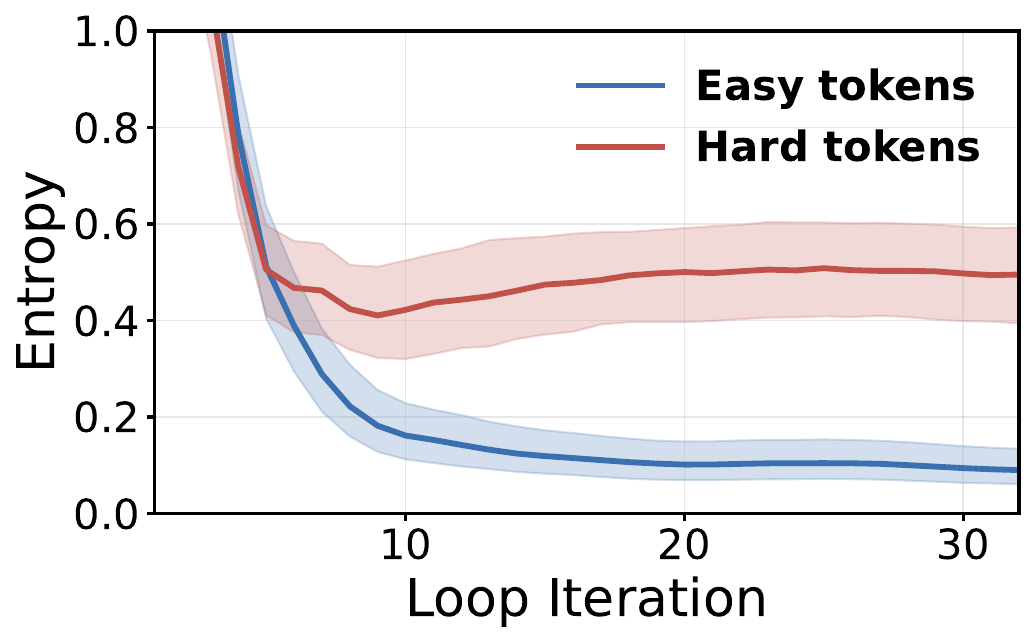}
        \caption{}
        \label{fig:main_c}
    \end{subfigure}

    \vspace{-2mm}
    \caption{
        (a) Histogram of trace entropy $H_{\mathrm{tr}}$ for correct and incorrect GSM8K reasoning trajectories. 
        (b, c) Logit margin and token entropy trajectories of easy and hard tokens across loop iterations.
    }
    \vspace{-4mm}
    \label{fig:main}
\end{figure*}

To mitigate this issue, we propose \textbf{LoopCD},\footnote{\url{https://github.com/hoeng4/LoopCD}} a training-free inference-time decoding method for LoopLMs that substantially improves reasoning performance. Our key idea is to exploit the internal dynamics exposed by looped computation: while later iterations provide more refined predictions, earlier iterations retain complementary intermediate signals useful for contrast. By leveraging this loop-wise discrepancy, we construct the final sampling distribution by contrasting logits from earlier iterations with those from the final refined iteration. Our method requires neither auxiliary amateur models nor architectural modifications, and adds negligible computational overhead by reusing hidden states already produced during looping. Moreover, it naturally refines reasoning-critical hard tokens rather than the entire trajectory, achieving reliable reasoning. 
Extensive experiments show that LoopCD consistently improves LoopLM reasoning across arithmetic reasoning, code generation, and question answering benchmarks, outperforming standard and advanced decoding baselines.

\setlength{\abovedisplayskip}{6pt}
\setlength{\belowdisplayskip}{6pt}

\vspace{-0.15cm}
\section{Preliminaries}

\textbf{Looped Transformer.}
A Looped Transformer applies recurrence along the depth dimension by repeatedly reusing a fixed stack of Transformer blocks. We denote its architecture by
$(p, k \otimes d, l)$, where $p$ and $l$ denote the numbers of layers in \textit{prelude} and \textit{coda}, corresponding to the layer stacks before and after the loop. $k$ denotes the number of recurrent layers, and $d$ is the recurrence depth.
Let $S_x$ denote an $x$-layer Transformer stack. $S'(\cdot)$ denotes untied Transformer stacks, while $S(\cdot)$ denotes
weight-tied stacks across recurrences. Given input hidden states $X$, a simplified forward computation of Looped Transformer is 
\begin{equation}
    H = S'_l\Bigl(
        \Bigl(
        \underbrace{S_k \circ S_k \circ \cdots \circ S_k}_{d\ \text{times}}
        \Bigr)(S'_p(X))
    \Bigr).
\end{equation}
Thus, the model has parameter depth $p+k+l$ but realized computational depth $p+kd+l$.
Finally, logits are produced after the last recurrence:
$
    p(x_{i+1}\mid x_{\leq i})
    =
    \mathrm{softmax}(W_{\mathrm{out}} H_i).
$
Unless otherwise specified, verbalization is performed at a fixed maximum recurrence depth $d_{\max}$ shared by all tokens. We mainly consider the case of $(2,4 \otimes32,2)$ Huginn-0125 \cite{huginn}.

\section{Motivation}

Recent works on evaluating the reasoning performance of large language models have shown that reasoning quality can be effectively estimated using the model's internal signals \cite{DeepThink,mti, kang2026scalable}. We also observe that the same approach can be safely applied to evaluating the reasoning performance of LoopLMs. We first define \textit{trace entropy} $H_{tr}$ to measure the model's uncertainty along the reasoning trajectory.
\begin{equation*}
H_{tr} = \frac{1}{N}\sum_{i=1}^{N}{H_i} \quad, H_i = -\sum_{j}p_i(j)\log\ p_i(j)
\end{equation*}
where $i$ denotes the token position. In Fig. \ref{fig:main}(a), we plot the histograms of $H_{tr}$ for correct and incorrect reasoning trajectories on the arithmetic reasoning task GSM8K. As shown, incorrect reasoning trajectories generally exhibit higher entropy, whereas correct trajectories tend to have lower entropy. This suggests that, although LoopLMs are implicitly trained to reason in latent space, they can still produce uncertain tokens during verbalization, which substantially affects reasoning performance.

This naturally raises the question: which tokens in the decoding process of LoopLMs give rise to such uncertainty? To investigate this, we use logit-lens \cite{nostalgebraist2020logitlens} to examine how tokens are refined across different recurrence iterations. We observe two distinct patterns. As shown in Fig. \ref{fig:main}(b), most tokens exhibit monotonic improvement behavior, where the margin between the top-1 and top-2 predictions increases over iterations. However, a minority subset of tokens exhibits the opposite behavior: the model becomes confident too early and later attempts to revise its initially confident prediction. We refer to the former case as \textit{easy} tokens and the latter as \textit{hard} tokens (formal definitions in \Cref{sec:definition_hard_tokens}). As shown in Fig. \ref{fig:main}(c), these hard tokens usually result in high-entropy distributions, and they account for about $10\%$--$15\%$ of the reasoning trace (Appendix \Cref{fig:hard_token_ratio}).

We observe that these hard tokens have a strong impact on reasoning quality. (1) As shown in Fig. \ref{fig:main}(c), hard tokens generally lead to higher entropy and therefore tend to contribute more to incorrect reasoning trajectories, as reflected in Fig. \ref{fig:main}(a). (2) 
We further investigate the impact of hard tokens through random token replacement. As shown in \Cref{tab:random_replacement_drop}, replacing hard tokens drops accuracy more substantially than replacing easy tokens, suggesting that the generated outputs are more sensitive to perturbations of hard tokens. This trend is further supported by the selective-intervention analysis in \Cref{subsec:ablation}, where LoopCD’s gains are concentrated on hard tokens.

\begin{table}[t]
\centering
\small
\begin{tabular}{lcc}
\toprule
\textbf{Correct$\rightarrow$Incorrect} & \textbf{Easy Tokens} & \textbf{Hard Tokens} \\

\midrule
GSM8K& 31.3\% & \textbf{52.9\%} \\
StrategyQA & 25.9\% & \textbf{71.9\%}  \\
\bottomrule
\end{tabular}
\vspace{-3mm}
\caption{Percentage of originally correct trajectories that become incorrect
after token replacement; accuracy is strict-match. Protocol details in \Cref{app:table1}.}

\label{tab:random_replacement_drop}
\vspace{-0.4cm}
\end{table}

\section{Method: LoopCD}

Motivated by the observations above, we propose \textbf{LoopCD} to efficiently modulate the behavior of hard tokens during LoopLM inference. Our method is based on contrastive decoding (CD)~\cite{cd}, but instead of using an external amateur model, it constructs the contrast internally from different iterations of looping. Specifically, at decoding step $t$, let $p_t^{(\ell)}(v)$ denote the probability assigned to token $v \in \mathcal{V}$ at loop iteration $\ell$. We treat the distribution from a final iteration $\ell_e$ as the \emph{expert} distribution, and the distribution from an earlier iteration $\ell_a < \ell_e$ as the \emph{amateur} distribution. We compute a contrastive score as:
\begin{align}
    s_t(v)
    =
    \log p_t^{(\ell_e)}(v)
    -
    \lambda \log p_t^{(\ell_a)}(v),
\end{align}
and select the next token from the normalized distribution
$
    p_{\mathrm{}}(v)
    =
    \frac{\exp(s_t(v))}
    {\sum_{u \in \mathcal{V}} \exp(s_t(u))}.
$
Here, $\lambda$ is the contrastive coefficient, controlling how strongly the amateur distribution is penalized.
We empirically choose the amateur loop index $\ell_a$, as discussed in more detail in \Cref{subsec:amaeur_loop}. To avoid promoting implausible tokens through contrastive scoring, we additionally apply the adaptive plausibility constraint described in \Cref{sec:implementation}.

\paragraph{Mechanism.} We now explain how our LoopCD improves reasoning by intervening on hard tokens from two perspectives.
\textbf{i) } It serves as a corrective mechanism for early-overconfident predictions. If an earlier iteration assigns high probability to an incorrect token, subtracting the early amateur log probability suppresses this premature preference and allows the later distribution to recover alternative tokens. \textbf{ii) }  It can be viewed as a form of \textit{contrasting guidance}, analogous to guidance commonly used in image generation. For example, diffusion models \cite{ho2020denoising} often construct a score of the form $
    (1+\alpha)\epsilon_{\theta}(x)
    -
    \alpha \epsilon_{\mathrm{bad}}(x),
$
where $\epsilon_{\mathrm{bad}}$ represents an undesirable or less informative score, 
such as an unconditional score or under-trained model~\cite{cfg, ag}. In LoopCD, the earlier loop iteration can naturally play the role of this less refined ``bad'' predictor. 
We provide further empirical evidence supporting this mechanism in \Cref{appendix:mechanism}.

\paragraph{LoopCD primarily affects hard tokens.}
One direct approach would be to explicitly identify hard tokens  and selectively intervene on them during decoding, like MTI \cite{mti}. However, this requires tracking token confidence during iteration,  and introduces an additional hyperparameter for \textit{hardness} threshold. Here, we show that our LoopCD can naturally avoid these issues by  having a stronger effect on confusing hard tokens.

Consider two candidate tokens $y$ and $z$ at decoding position $t$, and define the margin at loop iteration $\ell$ as
$ \Delta^{(\ell)}(y,z)
    =
    \log p_t^{(\ell)}(y)
    -
    \log p_t^{(\ell)}(z).$
Under LoopCD, the margin becomes:
\begin{align*}
    (1-\lambda)\underbrace{\Delta^{(\ell_e)}(y,z)}_{M_e}
    +
    \lambda
    \underbrace{
    \bigl(
    \Delta^{(\ell_e)}(y,z)
    -
    \Delta^{(\ell_a)}(y,z)
    \bigr)
    }_{M_{CD}}.
\end{align*}
For tokens exhibiting the typical easy token behavior in Fig. \ref{fig:main}(b), the margin increases from the amateur to the expert iteration, yielding $M_{CD} >0$, so LoopCD preserves the top-1 prediction. 
In contrast, hard tokens exhibit larger changes between the amateur and expert iterations due to peak-drop confidence dynamics, which can yield $(1-\lambda)M_{e} +\lambda M_{CD}<0 $ and change the top-1 prediction.
Our contrastive term is naturally much more influential for these hard tokens, allowing LoopCD to naturally refine hard tokens without explicitly detecting hard tokens or introducing an additional hyperparameter. We further investigate this implicit selectivity in \Cref{subsec:ablation}.

\begin{table*}[t]
\centering
\setlength{\tabcolsep}{4.5pt}
\renewcommand{\arraystretch}{0.9}
\small
\begin{tabular*}{\textwidth}{@{\extracolsep{\fill}}llcccccccccc}
\toprule
\multirow{2}{*}{\textbf{Model}} 
& \multirow{2}{*}{\textbf{Method}}
& \multicolumn{2}{c}{\textbf{GSM8K}}
& \multicolumn{2}{c}{\textbf{MATH-500}}
& \multicolumn{2}{c}{\textbf{HumanEval}}
& \multicolumn{2}{c}{\textbf{MBPP}}
& \multirow{2}{*}{\textbf{StrQA}}
& \multirow{2}{*}{\shortstack{\textbf{Rel. Tok/s}}}\\
\cmidrule(lr){3-4}
\cmidrule(lr){5-6}
\cmidrule(lr){7-8}
\cmidrule(lr){9-10}
& 
& \textbf{Flex} & \textbf{Strict}
& \textbf{Flex} & \textbf{Strict}
& \textbf{HE} & \textbf{HE+}
& \textbf{MBPP} & \textbf{MBPP+}
& & \\
\midrule

\rowcolor{gray!15}
\cellcolor{white}\multirow{6}{*}{\textbf{Huginn-0125}}
 &  Greedy
        & 33.21
        & 23.12
        & 13.60 
        & 12.20
        & 24.39 
        & 20.73
        & 40.74 
        & 33.60 
        & 54.02
        & $1.00\times$
        \\
 & MTI
        & 33.51
        & 24.11
        & 15.20
        & 13.40
        & 25.00
        & 20.73
        & 42.06
        & 35.71
        & 54.19
        &0.93$\times$
        \\ 
        
& Self-Eval
& 34.04
& 24.03
& 13.80
& 12.60
& 21.95
& 19.51
& 40.48
& 33.60
& 53.62
& 0.28$\times$
        \\
 & NoiseCD
        & 33.36
        & 22.06
        & 14.20
        & 13.40
        & 23.78
        & 20.73
        & \textbf{42.59}
        & 35.45
        & 53.14
        &0.98$\times$
        \\

\cmidrule(lr){2-12}
& \textbf{Ours}  
        & \textbf{36.62} 
        & \textbf{26.46}
        & \textbf{15.60} 
        & \textbf{15.20}
        & \textbf{30.49} 
        & \textbf{27.44} 
        & 42.06
        & \textbf{36.77} 
        & \textbf{55.24}
        & 0.98$\times$
\\
&    
        & \textcolor{green!50!black}{\textbf{+3.41}}  
        & \textcolor{green!50!black}{\textbf{+3.34}}
        & \textcolor{green!50!black}{\textbf{+2.00}}  
        & \textcolor{green!50!black}{\textbf{+3.00}}
        & \textcolor{green!50!black}{\textbf{+6.10}} 
        & \textcolor{green!50!black}{\textbf{+6.71}}  
        & \textcolor{green!50!black}{\textbf{+1.32}}  
        & \textcolor{green!50!black}{\textbf{+3.17}}  
        & \textcolor{green!50!black}{\textbf{+1.22}}  

\\

\midrule

\rowcolor{gray!15}  
\cellcolor{white}\multirow{6}{*}{\textbf{Ouro-1.4B}}
&  Greedy 
        & 78.70
        & 60.65
        & 50.20
        & 34.60
        & 69.50
        & 65.85
        & 72.75
        & 61.90
        & 64.00
        & 1.00$\times$
        \\
 & MTI
        & 76.88
        & 57.85
        & \textbf{54.20}
        & 22.60
        & 70.73
        & 65.24
        & 74.07
        & 62.96
        & 65.07
        & 0.93$\times$
        \\
 & Self-Eval
        & 77.06
        & 54.66
        & 46.80
        & 27.80
        & 74.39
        & 69.51
        & 69.31
        & 58.73
        & 56.86
        & 0.25$\times$
        \\
 & NoiseCD
        & 77.03
        & 57.92
        & 51.00
        & 34.20
        & 71.95
        & 67.68
        & 71.16
        & 59.26
        & 64.72
        & 0.99$\times$
        \\
\cmidrule(lr){2-12}
& \textbf{Ours}  
        & \textbf{81.05} 
        & \textbf{64.06}
        & 51.40 
        & \textbf{37.40}
        & \textbf{74.39} 
        & \textbf{70.12} 
        & \textbf{74.60} 
        & \textbf{62.96} 
        & \textbf{66.81}
        &0.99$\times$
\\
&    
        & \textcolor{green!50!black}{\textbf{+2.35}}  
        & \textcolor{green!50!black}{\textbf{+3.41}}  
        & \textcolor{green!50!black}{\textbf{+1.20}}  
        & \textcolor{green!50!black}{\textbf{+2.80}}  
        & \textcolor{green!50!black}{\textbf{+4.89}} 
        & \textcolor{green!50!black}{\textbf{+4.27}}  
        & \textcolor{green!50!black}{\textbf{+1.85}}  
        & \textcolor{green!50!black}{\textbf{+1.06}}  
        & \textcolor{green!50!black}{\textbf{+2.81}}  

\\
\bottomrule
\end{tabular*} \vspace{-2mm}
\caption{Performance and inference-speed comparison for Huginn and Ouro. Tokens/s is measured on MATH-500. 
Flex/Strict: flexible/canonical matching; ``+'': EvalPlus. See \Cref{sec:appendix_benchmark} for details.}
\vspace{-4mm}
\label{tab:MainTable}
\end{table*}

\section{Experiments}

\noindent\textbf{Models. }
We focus our experiments on two different pretrained LoopLMs: Huginn-0125~\cite{huginn} and Ouro-1.4B~\cite{ouro}, which was trained with substantially more compute and on a larger dataset.

\noindent\textbf{Benchmarks. }
We evaluate our method on three reasoning domains: (1) Mathematical reasoning on GSM8K \cite{gsm8k} and MATH-500 \cite{math500}, (2) Code generation on HumanEval \cite{humaneval} and MBPP \cite{mbpp}, and (3) commonsense QA on StrategyQA \cite{strategyqa}. Further details are provided in \Cref{sec:appendix_benchmark}.

\noindent\textbf{Baselines. }
We primarily compare our method with \textit{greedy decoding}, which corresponds to the standard decoding of LoopLMs. We further compare our method with several advanced decoding baselines: 
\textbf{(a) MTI}~\cite{mti} is the most closely related baseline to ours, which intervenes on high-entropy tokens using negative-prompted classifier-free guidance.
\textbf{(b) Self-Eval } \cite{selfeval} samples multiple candidate reasoning trajectories and selects the final output through self-evaluation-based reranking, at the cost of increased computation.
\textbf{(c) NoiseCD. } We also test a CD variant that injects noise into final-loop hidden states and contrasts the resulting logits, following common CD strategies in VLMs~\cite{VLMCD}. 

\noindent\textbf{Implementation Details. }
LoopCD has two hyperparameters: CD coefficient $\lambda$ and the amateur loop index $\ell_a$. The expert loop index $\ell_e$ is fixed to the final recurrent iteration of each model, i.e., 32 for Huginn-0125 and 4 for Ouro-1.4B. For a fair comparison, we choose each baseline using the same protocol and report its best setting. Further details are provided in \Cref{sec:implementation,sec:hyperparameters}.

\subsection{Main Results}
As shown in \Cref{tab:MainTable}, LoopCD shows the best overall performance compared to standard greedy decoding and other decoding variants. 
We summarize the main observations below: \textbf{(a) MTI. } Although MTI can improve over greedy decoding, it requires a sensitive entropy-threshold hyperparameter and an additional negative-condition KV-cache pass, reducing throughput. In contrast, LoopCD achieves better performance without such extra components, suggesting that earlier-loop logits provide a more suitable contrastive signal for LoopLMs. 
\textbf{(b) Self-Eval. } Compared with this method based on multiple parallel rollouts, which substantially reduces throughput, LoopCD achieves strong reasoning performance while remaining efficient. 
\textbf{(c)  NoiseCD. } NoiseCD shows improvement in only limited cases. Unlike LoopCD, it also affects easy tokens due to hidden-state noise injection, which can limit its performance gains. This supports our motivation for intervening on reasoning-critical hard tokens and shows the effectiveness of logit-space contrasting.

\begin{table}[t]
\centering
\small
\begin{tabular}{l c c c}
\toprule
Method & GSM8K & MATH-500 & Tok/s\\
\midrule
\rowcolor{gray!15} Greedy
        & 33.21 
        & 13.60 
        & 5.09 (1.00$\times$)\\
\textbf{Ours}
        & \textbf{36.62}
        & \textbf{15.60}
        & 4.99 (0.98$\times$)\\
EE
         & 33.66
         & 14.60
         & 10.06 (1.98$\times$)\\
EE+\textbf{Ours}
         & 34.12
         & \textbf{15.60}
         & 9.89 (1.94$\times$)\\
\bottomrule
\end{tabular} 

\vspace{-2mm}
\caption{
Flex accuracy and throughput of LoopCD with Early Exit (EE) on Huginn.
}
\vspace{-2mm}
\label{tab:inference_latency}
\end{table}

\begin{table}[t]
\centering
\small
\begin{tabular}{l c c c c c}
\toprule
 Loop Idx. & 4 & 8 & 12 & 16 & 20\\
\midrule
GSM8K & 35.41 & \textbf{36.39} & 34.42 & 33.51 & 32.90 \\
HumanEval & 26.83 & \textbf{30.49} & 26.22 & 24.39 & 25.00 \\

\bottomrule
\end{tabular} 

\vspace{-2mm}
\caption{Amateur loop   sensitivity on Huginn ($\lambda=0.3$).}
\vspace{-4mm}
\label{tab:amateur}
\end{table}

\begin{figure*}[t]
    \centering
    \includegraphics[width=1.0\linewidth]{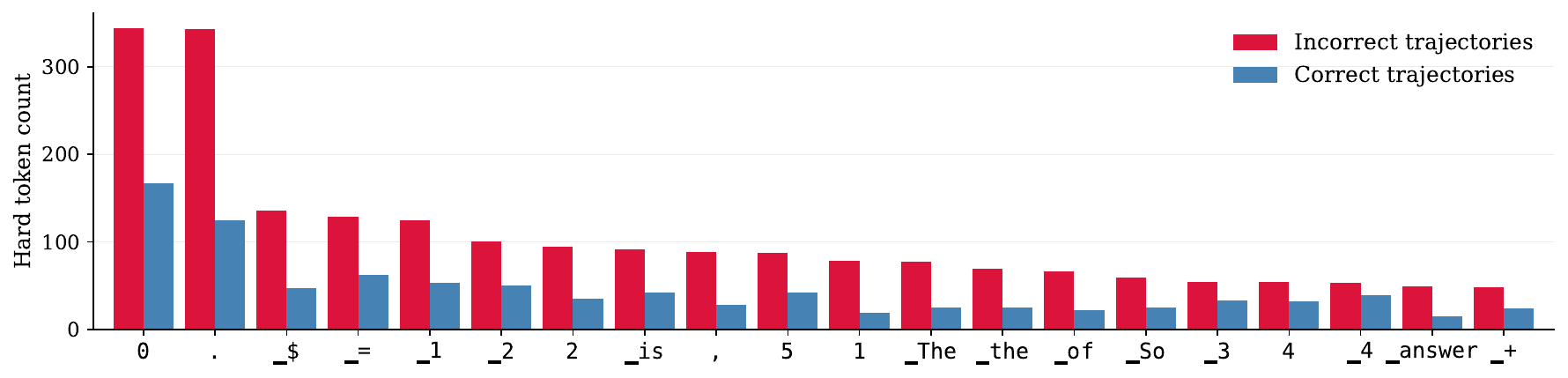}
    \vspace{-8mm}
    \caption{
    Top-20 hard token types from Huginn-0125 greedy generations on 600 GSM8K examples.
    Numerical tokens, arithmetic operators, punctuation, and reasoning-step markers appear frequently, suggesting that hard tokens are concentrated on tokens that affect intermediate computations and reasoning structure.
    }
    \vspace{-4mm}

    \label{fig:ablation_top_20}
\end{figure*}

\subsection{Compatibility with Early Exit}
We also combine LoopCD with early exit, which stops recurrence before the final loop when the hidden states change only marginally across consecutive iterations.
As shown in \Cref{tab:inference_latency}, early exit substantially improves throughput by reducing recurrent computation. 
LoopCD can be combined with early exit to improve reasoning quality while retaining the efficiency gains of early exit.

\subsection{{Amateur Loop Index Sensitivity}}
\label{subsec:amaeur_loop}

We study the sensitivity to the amateur loop index in LoopCD.
As shown in \Cref{tab:amateur}, choosing an early loop, roughly 15--30\% of the recurrence depth, often works well across tasks and models.
This suggests that early representations are predictive enough while remaining distinct from the final iteration. 
We further analyze why this intermediate amateur range is particularly effective in \Cref{appendix:amateur_iteration}.

\subsection{Ablation: Where Do LoopCD's Gains Come From?}
\label{subsec:ablation}
Due to their non-monotonic loop-wise dynamics, hard tokens are expected to be more strongly affected by LoopCD. 
We empirically examine this implicit selectivity and whether it accounts for LoopCD's performance gains.

We first examine what kinds of tokens exhibit hard dynamics during reasoning.
As shown in \Cref{fig:ablation_top_20}, hard tokens frequently include numbers and arithmetic operators, which affect intermediate computations, as well as punctuation and reasoning-step markers such as \texttt{The} and \texttt{So}, which mark boundaries and transitions between reasoning steps. This suggests that loop instability is concentrated on tokens that shape intermediate quantities and the structure of the reasoning trajectory.

We next test whether hard tokens drive LoopCD's gains by selectively applying it to hard or easy tokens. 
As shown in \Cref{tab:selective_intervention}, hard-only intervention recovers much of the full gain despite affecting far fewer tokens, while easy-only intervention yields smaller improvements. This confirms that LoopCD's gains are concentrated on hard tokens. The same trend holds within numeric and operator tokens, where hard subsets consistently outperform their easy counterparts. Nevertheless, full LoopCD performs best, suggesting that explicitly restricting intervention to predefined token subsets can miss useful corrections. This supports our design of applying LoopCD to all tokens and relying on its implicit selectivity, rather than explicitly detecting hard tokens.

\begin{table}[t]
\centering
\small
\begin{tabular}{l c c c}
\toprule
Setting & Acc. (\%) & $\Delta$ & Interv. (\%) \\
\midrule
\rowcolor{gray!15} Greedy
        & 33.21
        & 
        & 0 \\

LoopCD (full)
        & 36.39
        & \textcolor{green!50!black}{\textbf{+3.18}}
        & 100 \\

\midrule

\textbf{Hard only}
        & \textbf{35.71}
        & \textcolor{green!50!black}{\textbf{+2.50}}
        & \textbf{10.7} \\

Easy only
        & 34.65
        & \textcolor{green!50!black}{\textbf{+1.44}}
        & 89.0 \\

\midrule 

\textbf{Hard numeric only}
        & \textbf{34.04}
        & \textcolor{green!50!black}{\textbf{+0.83}}
        & \textbf{2.9} \\

Easy numeric only
        & 32.52
        & \textcolor{red!70!black}{\textbf{-0.68}}
        & 26.2 \\

\midrule

\textbf{Hard operator only}
        & \textbf{33.51}
        & \textcolor{green!50!black}{\textbf{+0.30}}
        & \textbf{0.7} \\

Easy operator only
        & 33.28
        & \textcolor{green!50!black}{\textbf{+0.08}}
        & 6.9 \\
\bottomrule
\end{tabular}

\vspace{-2mm}
\caption{
Performance of LoopCD under selective intervention on hard/easy tokens and numeric/operator tokens on Huginn ($\ell_a=8,\lambda=0.3$).
}
\vspace{-4mm}
\label{tab:selective_intervention}
\end{table}

\subsection{Further Analysis}
\label{subsec:further}
We also present an extensive analysis of loop instability and the behavior of LoopCD (\Cref{sec:analysis}), along with qualitative studies (\Cref{sec:qualitative}).
To assess the robustness of LoopCD, we report results with fixed hyperparameter settings across benchmarks and multi-seed evaluations (\Cref{sec:robustness_loopcd}).

\section{Conclusion} 
In this work, we showed that LoopLM reasoning trajectories contain reasoning-critical hard tokens whose non-monotonic loop-wise dynamics can substantially affect reasoning outcomes. To address this, we proposed LoopCD, an efficient inference-time method that contrasts earlier-loop logits with final-loop logits to effectively intervene on such tokens. Extensive experiments show that our LoopCD can consistently improve reasoning performance without additional training or architectural changes across diverse reasoning tasks.

\newpage

\section*{Limitations}

One limitation of LoopCD is that it relies on a training-free heuristic for selecting the contrastive signal. While this makes the method simple and efficient, the chosen amateur loop may not always provide an optimal contrast for every token, task, or model. We expect further improvements to be achievable by learning a verifier that identifies more reliable contrastive signals, or by adaptively selecting the optimal contrasting loop index during decoding.

\section*{Acknowledgments}
This work was supported by IITP and NRF grant funded by the Korea government(MSIT) (RS-2026-25490269, RS-2019-II191906)


\bibliography{custom}

\newpage
\appendix
\crefalias{section}{appendix}
\crefalias{subsection}{appendix}

\section{Related Works}

\paragraph{Reasoning in Language Models.}

With the development of large language models, research interest has gradually shifted from simple text generation to whether LLMs can perform reasoning required for real problem solving. 
A growing line of CoT-based reasoning methods \cite{CoT,s1,cot2} has shown the effectiveness of verbalizing intermediate reasoning steps for improving LLM reasoning.
Following this line, later studies explored how models can learn to generate better CoT reasoning through training or reinforcement learning \cite{rlrl, liu2024deepseek}.

Another line of work has attempted to further improve reasoning through inference-time scaling. Tree of Thoughts \citep{yao2023tree, tree2} explores multiple reasoning trajectories and selects more plausible paths through evaluation. Recent studies have further investigated whether reasoning quality can be estimated without a separate reward model, using signals from the model itself. Self-Consistency \cite{selfconsistency} shows that sampling multiple reasoning paths and selecting answers based on their agreement can serve as an effective inference-time strategy. More recent work such as DeepThink \cite{DeepThink} shows that localized log-probability-based confidence within reasoning trajectories can effectively measure reasoning quality and enable more efficient test-time scaling.

\paragraph{Recurrence Models.}

Recurrence models, which repeatedly apply a fixed set of weights to refine representations in latent space, have been widely explored across many areas of machine learning. Early recurrent architectures, such as RNNs and LSTMs \citep{rnn, lstm}, achieved strong results in time-series modeling and natural language processing by repeatedly applying shared parameters over sequential inputs. Deep Equilibrium Models \cite{deq1,deq2} apply a shared transformation recurrently in latent space and directly solve for its fixed point, providing an implicit-depth architecture that can achieve competitive performance.

A major recent application of recurrence appears in generative modeling. Diffusion models \citep{ho2020denoising, ddim} generate data by iteratively applying a denoising network to progressively refine noisy variables, and have shown strong performance across image, text, and audio generation. More recently, instead of relying on the standard sequential stack of distinct neural network layers, looped or recursive thinking models have attracted attention by repeatedly applying the same layer or module weights \citep{hrm, trm, urm}. Recent studies find that such recurrent computation can achieve strong performance on reasoning tasks such as Sudoku and maze solving with far fewer parameters than conventional large neural networks.

\paragraph{Contrastive Decoding.}
Contrastive Decoding (CD) was originally proposed to improve the quality of open-ended text generation by contrasting an expert language model with a weaker amateur model, using their likelihood difference to reduce undesirable patterns such as repetition and incoherence~\citep{cd}.
Beyond its original setting, subsequent work has shown that contrastive decoding can improve reasoning~\citep{reasoningcontrastive} and mitigate hallucination~\citep{VLMCD}.
A separate line of work constructs contrastive views from the same model using intermediate or early exit layers, model perturbations, or input distortion ~\citep{dola,acd,prunecd,icd}.
These methods suggest that CD does not require a separate model, but rather an informative negative view of undesirable or premature predictions. Recent work further shows that such intervention need not be applied uniformly across tokens.
In particular, MTI \citep{mti} identifies high-entropy tokens as major sources of reasoning errors and applies test-time intervention only at such uncertain positions.

\section{Experimental Details}
\subsection{Implementation Details}
\label{sec:implementation}
\paragraph{Adaptive Plausibility Constraint.}
Contrastive decoding can assign a high score to tokens that have low probability under the amateur distribution, even when those tokens are unlikely under the expert distribution. To avoid selecting such implausible tokens, we follow prior work~\cite{cd} and adopt an Adaptive Plausibility Constraint.
\begin{align}
    &\mathcal{V}_{\mathrm{head}}(x_{<i}) = \\
    &\big\{
    x_i \in \mathcal{V}
    :
    p_{\mathrm{e}}(x_i \mid x_{<i})
    \ge \nonumber
    \alpha \max_{w \in \mathcal{V}} p_{\mathrm{e}}(w \mid x_{<i})
    \big\}.
    \label{plausability_constraint}
\end{align}
Only tokens in $\mathcal{V}_{\mathrm{head}}(x_{<i})$ are considered when applying the CD objective. In all experiments, we set $\alpha=0.1$, following the original setup. MTI and NoiseCD also employ the adaptive plausibility constraint with $\alpha=0.1$.

\paragraph{Deterministic Initialization.}
Huginn~\cite{huginn} begins recurrent computation from an initial recurrent state, which is randomly initialized from a Gaussian distribution in the original model. For deterministic evaluation, we instead set this initial state to zero in our main experiments. We further verify robustness to stochastic initialization across multiple random seeds in \Cref{appendix:multi_seed}.

\paragraph{MTI~\cite{mti}.}
MTI intervenes only at uncertain decoding positions, using the model itself
to construct a negative condition. At decoding step $t$, we take the
distribution obtained after the full recurrence as
$p_t^{\mathrm{cond}} = p_t^{(\ell_e)}$, and regard the position as uncertain
when its Shannon entropy exceeds a threshold,
$H_t > \tau$.
\begin{equation}
H_t
=
-\sum_{v \in \mathcal{V}}
p_t^{\mathrm{cond}}(v)
\log p_t^{\mathrm{cond}}(v).
\end{equation}

For such positions, we obtain an unconditional distribution by appending
a fixed negative prompt $c_{\mathrm{neg}}$ to the prefix, and
combine the two with classifier-free guidance:

\begin{equation}
s_t^{\mathrm{MTI}}(v)
=
\begin{cases}
s_t^{\mathrm{CFG}}(v), & H_t > \tau, \\
\log p_t^{\mathrm{cond}}(v), & \text{otherwise},
\end{cases}
\end{equation}

\begin{equation}
s_t^{\mathrm{CFG}}(v)
=
(1-\omega)\log p_t^{\mathrm{uc}}(v)
+\omega\log p_t^{\mathrm{cond}}(v).
\end{equation}
The next token is selected as $\arg\max_v s_t^{\mathrm{MTI}}(v)$. We use the original negative prompt $c_{\mathrm{neg}}=\texttt{OUTPUT ERROR}$. 

The official Hugging Face implementation deep-copies the KV cache before computing $p_t^{\mathrm{uc}}$, which is expensive for LoopLMs due to their recurrent cache. We instead apply the negative prompt directly to the cache and truncate the appended entries afterwards. This yields the same $p_t^{\mathrm{uc}}$ without cache copying, so each intervention requires only one additional negative-prompt forward pass.

\paragraph{Self-Evaluation ~\cite{selfeval}.}
For each problem $x_i$, we generate $N=4$ candidate answers 
$\{y_{i,k}\}_{k=1}^{N}$ and select one using self-evaluation. 
The candidate generation procedure differs by model. 
For Ouro, we use stochastic decoding with temperature. 
For Huginn, we keep token decoding greedy and rely on randomized initial recurrent states to produce different candidates, since temperature-based token sampling degraded performance in our preliminary trials.

We then score each candidate with the same deterministic self-evaluation procedure. 
Given a candidate answer $y_{i,k}$, we construct a prompt asking whether the answer is correct:
\begin{quote}
\small
Q: \{question\}\\
Answer: \{candidate answer\}\\
Is the above answer correct?\\
A) Yes\\
B) No\\
Answer:
\end{quote}
We run one deterministic forward pass and extract the logits of the single-token labels `` A'' and `` B'' at the final position. 
Let $z_A(y_{i,k})$ and $z_B(y_{i,k})$ denote these logits. 
We define the self-evaluation score as
\begin{equation}
s(y_{i,k}) = z_A(y_{i,k}) - z_B(y_{i,k}),
\end{equation}
and select the candidate with the highest score:
\begin{equation}
y_i^\star = y_{i,k^\star}, 
\quad
k^\star = \arg\max_{k} s(y_{i,k}).
\end{equation}

\paragraph{NoiseCD.}
To test whether the improvement of LoopCD comes from the structured recurrent
trajectory or merely from contrasting against a corrupted distribution, we introduce
a noisy-amateur contrastive decoding baseline. Instead of using an earlier loop
iteration as the amateur distribution, this baseline constructs an artificial amateur
by perturbing the final hidden state with Gaussian noise.

Let $h_t^{(\ell_e)} \in \mathbb{R}^{d}$ denote the final expert hidden state at decoding
position $t$. We construct a corrupted hidden state as
\begin{align}
    h_{t,\mathrm{noise}}
    &=
    h_t^{(\ell_e)}
    +
    \sigma \cdot \mathrm{RMS}\!\left(h_t^{(\ell_e)}\right)
    \cdot
    \epsilon_t
    \\
    \epsilon_t &\sim \mathcal{N}(0, I), \nonumber
\end{align}
where $\sigma$ controls the relative noise strength. The RMS is computed over the
hidden dimension:
\begin{align}
    \mathrm{RMS}(h)
    =
    \sqrt{\frac{1}{d}\sum_{j=1}^{d} h_j^2}.
\end{align}
We then obtain the noisy-amateur distribution by applying the same language-model
head to the corrupted hidden state:
\begin{align}
    p_t^{(\mathrm{noise})}(v)
    =
    \mathrm{softmax}\!\left(g_\theta(h_{t,\mathrm{noise}})\right)_v .
\end{align}
Finally, we apply the same contrastive decoding rule as LoopCD:
\begin{align}
    s_t^{\mathrm{NoiseCD}}(v)
    =
    \log p_t^{(\ell_e)}(v)
    -
    \lambda \log p_t^{(\mathrm{noise})}(v).
\end{align}

This baseline removes the temporal structure of the recurrent trajectory while
preserving a comparable corrupted-amateur setup. Therefore, comparing LoopCD
against NoiseCD isolates whether the benefit comes from the earlier
loop iteration itself, rather than from subtracting an arbitrary noisy distribution.

\paragraph{Early Exit.}
LoopLMs compute by repeatedly applying a recurrent block to refine latent hidden states. 
However, once the hidden state changes only marginally between consecutive recurrent steps, the remaining iterations may provide little additional computation for the current token. 
This observation motivates early exit, which terminates the recurrent loop when the latent representation has sufficiently stabilized. 

For Huginn, we follow the early exit criterion described in the original
paper~\cite{huginn}. At recurrent iteration $r$, we compute the normalized change of the
hidden state between two consecutive iterations:
\begin{equation}
d_r
=
\frac{1}{T}
\sum_{j=1}^{T}
\frac{
\|h_{r,j}-h_{r-1,j}\|_2
}{
\|h_{r,j}\|_2
}.
\end{equation}
We exit at the first recurrent iteration satisfying $d_r < \tau$. We set $\tau=0.03$. 
When combining early exit with LoopCD, we allow early exit only after 15 recurrent iterations and use the early exit iteration as the expert loop. We separately search the amateur loop index for the combined setting. 

\subsection{Hyperparameters}
\label{sec:hyperparameters}
We describe the hyperparameters considered for LoopCD and each baseline.
For fair comparison, we use the best-performing setting for each method in
our reported results.

\paragraph{LoopCD.}
LoopCD has two hyperparameters: the amateur loop index $\ell_a$ and the CD coefficient $\lambda$.
The amateur loop index selects the intermediate loop output used as the amateur distribution.
Very early iterations can be noisy, while later iterations can be too close to the expert distribution, weakening the contrastive signal.
We therefore search $\ell_a$ over $\{6,8,10\}$ for Huginn and $\{0,1\}$ for Ouro.
For the contrastive strength, we search $\lambda \in \{0.1,0.2,0.3\}$ for both models.

\paragraph{MTI.}
MTI has two hyperparameters: the guidance scale and the entropy threshold.
For the guidance scale, we follow the original MTI setting and set it to $1.5$.
Since the entropy threshold can vary across models and benchmarks, we search it over $\{0.5, 1.0, 1.5\}$ for both Huginn and Ouro.

\paragraph{Self-Evaluation.}
For Ouro, we use temperature sampling with nucleus sampling.
In addition to the original Ouro decoding setting
($T=1.0$, $p=0.7$), we also evaluate ($T=0.7$, $p=0.9$) and report the
better performing configuration.

\paragraph{NoiseCD.}
NoiseCD uses the same contrastive coefficient $\lambda$ as LoopCD, and we fix
$\lambda=0.3$ in all experiments. Its additional hyperparameter is the
noise scale $\sigma$, which controls the strength of the Gaussian
perturbation added to the expert hidden state. We set $\sigma=0.5$ as a perturbation level.

\subsection{Benchmark Details}
\label{sec:appendix_benchmark}
\paragraph{GSM8K.}
We evaluate on the official GSM8K test split of 1,319 problems using the 3-shot prompt.
We report both Strict and Flex accuracy. Strict requires the canonical final-answer format, while Flex extracts the final numerical answer from the completion.

\paragraph{MATH-500.}
We evaluate on MATH-500, a 500-problem subset of MATH~\cite{math500}, introduced by \citet{math500_openai}, using the 4-shot Minerva prompt~\cite{minerva}.
We report Strict accuracy with the Minerva-MATH evaluator and Flex accuracy with \texttt{math\_verify}, which accepts mathematically equivalent answer forms.

\paragraph{HumanEval / HumanEval+.}
We evaluate HumanEval, a Python code-generation benchmark consisting of 164 function-completion problems, and HumanEval+, its EvalPlus-augmented version~\cite{evalplus} with additional test cases.
We use zero-shot greedy decoding with the original function signature and docstring as the prompt.
Following EvalPlus, we report both base-test and plus-test results.
In our tables, \textbf{HE} and \textbf{HE+} denote the base and plus test results, respectively.

\paragraph{MBPP / MBPP+.}
We evaluate MBPP, a Python code-generation benchmark consisting of short natural-language programming tasks, and MBPP+, its EvalPlus-augmented version with additional test cases.
We use the sanitized 378-problem MBPP+ split and perform zero-shot greedy decoding.
Following EvalPlus, we report both base-test and plus-test results.
In our tables, \textbf{MBPP} and \textbf{MBPP+} denote the base and plus test results, respectively.

\paragraph{StrategyQA.}
We evaluate StrategyQA, a binary commonsense QA benchmark requiring implicit multi-step reasoning.
We use the 2,290-question training split because the official test labels are hidden.
We use a 6-shot Chain-of-Thought prompt and greedy decoding.
We extract the final \texttt{yes}/\texttt{no} prediction and report accuracy.

\section{Formal Definition of Hard Tokens}
\label{sec:definition_hard_tokens}
For our Huginn-0125 analyses, we define hard tokens as follows.
For each generated position, let $z_i(v)$ denote the logit assigned to token
$v$ at recurrent iteration $i$, and let
\[
    y^\star = \arg\max_{v \in \mathcal{V}} z_{32}(v)
\]
be the final winner token at the last recurrent iteration. We define the
fixed-target logit margin of $y^\star$ at iteration $i$ as
\[
    m_i
    =
    z_i(y^\star)
    -
    \max_{v \in \mathcal{V},\, v \neq y^\star} z_i(v).
\]
The peak margin and final margin are then
\[
    m_{\mathrm{peak}} = \max_i m_i,
    \qquad
    m_{\mathrm{final}} = m_{32}.
\]
We define the confidence drop of the final winner as
\[
    d = m_{\mathrm{peak}} - m_{\mathrm{final}}.
\]
A generated token is classified as hard if
\[
    d \geq \epsilon,
\]
and as easy otherwise. 
We set $\epsilon=1.5$ throughout our experiments, under which approximately $89.3\%$ of generated tokens are classified as easy.

\section{Token replacement protocol}
\label{app:table1}
For each example correctly answered under greedy decoding, we sample
\(k=\min(5, |\textsc{hard}|, |\textsc{easy}|)\) positions from either hard or easy tokens and replace each with a token sampled uniformly from the top-3 candidates at that position, while keeping the rest of the completion fixed. We use the same replacement budget for both classes within each example to ensure a comparable setting. Examples lacking either class are excluded, leaving \(n=303\) for GSM8K and \(n=551\) for StrategyQA.

We evaluate the perturbed completions using strict-match accuracy based on the benchmark-specific answer format. Under this protocol, perturbing hard tokens causes a substantially larger accuracy drop than perturbing easy tokens, showing that the generated outputs are more sensitive to perturbations of hard tokens. This trend is further supported by the selective-intervention analysis in \Cref{subsec:ablation}, where LoopCD’s gains are concentrated on hard tokens.

\section{Analysis of Loop Instability and LoopCD Behavior}
\label{sec:analysis}

We provide additional analyses to complement the ablation in \Cref{subsec:ablation} and better understand why LoopCD's gains are concentrated on hard tokens. We first examine whether hard tokens are more prevalent in incorrect reasoning trajectories, then quantify whether LoopCD naturally affects hard tokens more strongly than easy tokens, and finally analyze why early recurrent iterations provide effective amateur distributions.

\subsection{Finding 1: Hard Tokens Are More Prevalent in Incorrect Trajectories}

\begin{figure}[t]
    \centering    \includegraphics[width=1.0\linewidth]{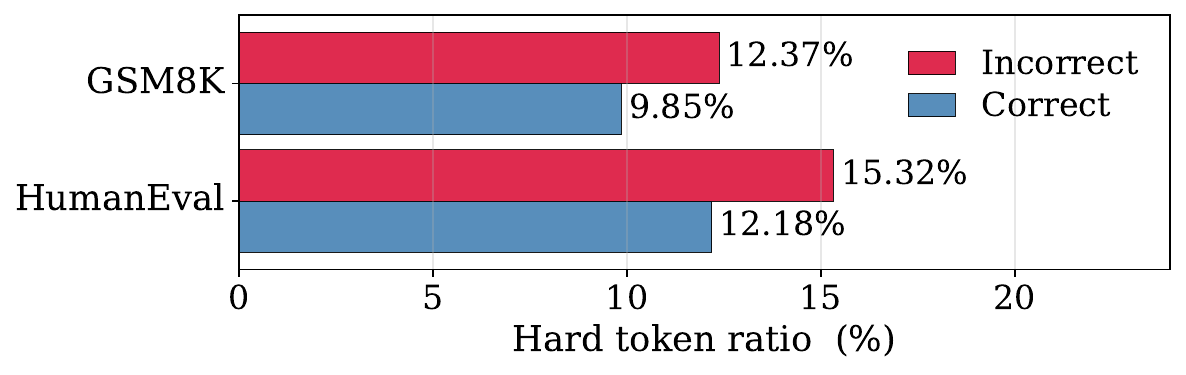}
    \vspace{-8mm}
    \caption{Proportion of hard tokens in correct and incorrect reasoning trajectories. Hard tokens account for roughly 10-15\% of generated tokens and are more prevalent in incorrect trajectories.
    }
    \vspace{-2mm}
    \label{fig:hard_token_ratio}
\end{figure}

We first examine whether hard tokens are associated with unsuccessful reasoning trajectories. As shown in Figure~\ref{fig:hard_token_ratio}, incorrect generations consistently contain a higher proportion of hard tokens than correct generations on both GSM8K and HumanEval. This suggests that loop instability is more prevalent in failed reasoning trajectories, further highlighting the importance of hard tokens in reasoning failures.

\begin{table}[t]
\centering
\small
\begin{tabular}{lccc}
\toprule
\textbf{Metric} & \textbf{Easy Tokens} & \textbf{Hard Tokens} & \textbf{Ratio} \\
\midrule
Margin Flip & 0.82\% & 8.78\% & 10.7$\times$ \\
\bottomrule
\end{tabular}
\vspace{-2mm}
\caption{
Margin-flip rates for easy and hard tokens on Huginn with GSM8K ($\ell_a=8$, $\lambda=0.3$).
}
\vspace{-4mm}

\label{tab:margin_flip_ratio}
\end{table}

\subsection{Finding 2: LoopCD Disproportionately Affects Hard Tokens}
\label{appendix:finding_2}
Given that hard tokens are more prevalent in wrong reasoning trajectories, we next examine whether LoopCD naturally affects these tokens more strongly than easy tokens. 
For each generated token $t$, let
\[
    y_t^{\star} = \arg\max_v z_{32,t}(v)
\]
be the greedy prediction at the final recurrent iteration. 
Given an amateur iteration $\ell_a$ and contrastive weight $\lambda$, we define the LoopCD contrastive logit as
\[
    \tilde z_t(v) = z_{32,t}(v) - \lambda z_{\ell_a,t}(v).
\]
We say that token $t$ is \textit{margin-flipped} if the contrastive logit changes the original greedy prediction:
\[
    \tilde z_t(y_t^{\star})
    <
    \max_{v \neq y_t^{\star}} \tilde z_t(v).
\]
This diagnostic measures whether the LoopCD contrast is strong enough to alter the final greedy decision without explicitly identifying hard tokens during decoding.

\paragraph{Result.}
As shown in Table~\ref{tab:margin_flip_ratio}, among 58,447 generated tokens,
hard tokens ($N=6{,}244$) are margin-flipped at a rate of 8.78\%,
whereas easy tokens ($N=52{,}203$) are flipped at only 0.82\%.
This corresponds to a 10.7$\times$ higher margin-flip rate for hard
tokens, showing that although LoopCD is applied to all tokens, its
contrastive effect naturally concentrates on hard tokens.

\begin{figure}[t]
    \centering
    \includegraphics[width=1.0\linewidth]{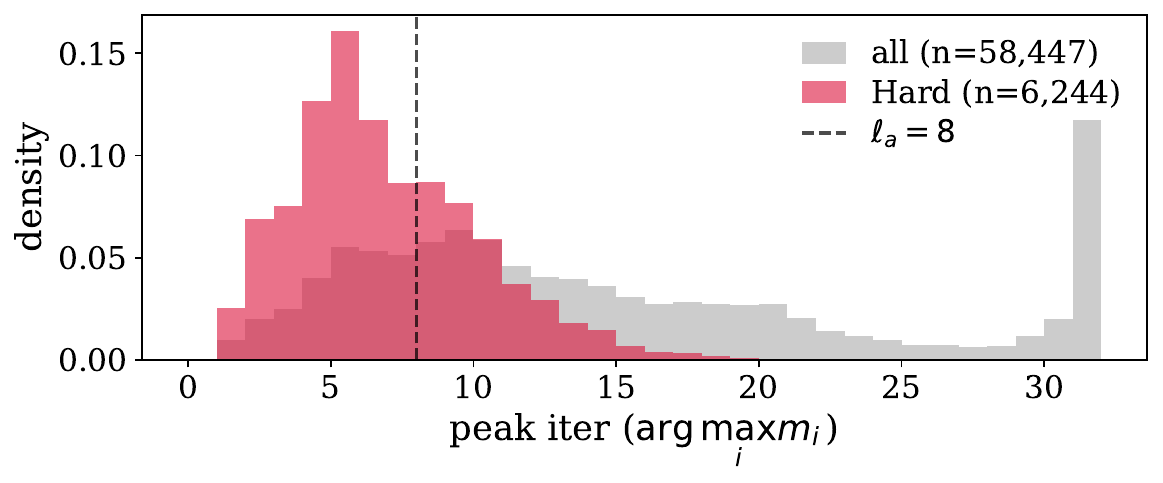}
    \vspace{-8mm}
    \caption{
    Peak iteration of the fixed-target margin $\arg\max_i m_{t,i}$ for hard and all tokens. 
    }
    \vspace{-2mm}
    \label{fig:appendix_peak_iter}
\end{figure}

\subsection{Finding 3: Hard Tokens Peak Near the Amateur Iteration}
\label{appendix:amateur_iteration}
We next investigate why an intermediate recurrent iteration, around 15--30\% of the recurrence depth, provides an effective amateur distribution. 
For each generated token $t$, we track the fixed-target margin of the final greedy prediction across recurrent iterations:
\[
    m_{t,i}
    =
    z_{i,t}(y_t^{\star})
    -
    \max_{v \neq y_t^{\star}} z_{i,t}(v),
\]
and record the iteration where this margin is maximized,
$\arg\max_i m_{t,i}$.

As shown in \Cref{fig:appendix_peak_iter}, hard tokens concentrate sharply around iterations 6-12, with a mode near $\ell_a=8$. In contrast, the full token population, which is dominated by easy tokens, is more broadly distributed and shifts toward later iterations. 
This suggests that $\ell_a=8$ is not merely an arbitrary hyperparameter: it captures the transient overconfidence of hard tokens, where the amateur-expert contrast becomes most effective. 
This observation is also consistent with the performance sweep in the main text, where amateur iterations around $\ell_a \approx 8$ work best.

\section{Expert/Amateur Mechanism and Failure Mode Distinction}
\label{appendix:mechanism}

\subsection{Justification of the Final-Loop Expert}

In LoopCD, ``expert'' and ``amateur'' are relative terms inherited from
contrastive decoding and do not imply guaranteed correctness. We use the
final iteration as the expert because it is the distribution from which
the original LoopLM verbalizes, while an earlier iteration provides the
contrastive signal. LoopCD therefore amplifies the early-to-final update
direction, which improves reasoning on average, but may reinforce an
error when later iterations move away from a better earlier prediction.

We distinguish this failure mode from early overconfidence by analyzing Huginn-0125 on GSM8K. We define a \textit{Late-Flip} position as one where the final-loop winner is not top-1 at least once during the latter half of the recurrent iterations. We define an early-overconfidence position as one satisfying the same confidence-drop criterion used to define hard tokens in \Cref{sec:definition_hard_tokens}, while its final-loop winner remains top-1 throughout the latter half of the recurrence (i.e., it is not a Late-Flip). Since a Late-Flip can be beneficial or harmful, its frequency provides an upper bound on potential late-loop erroneous revisions.

\begin{table}[t]
\centering
\small
\begin{tabular}{lcc}
\toprule
Type & Correct & Wrong \\
\midrule
Early Overconfidence & 9.4\% & 11.1\% \\
Late-Flip & 2.45\% & 2.96\% \\
\bottomrule
\end{tabular}
\caption{Frequency of early overconfidence and Late-Flip positions in
Huginn GSM8K greedy generations.}
\label{tab:late_flip_frequency}
\end{table}

As shown in \Cref{tab:late_flip_frequency}, Late-Flips are substantially less frequent than early-overconfidence positions, occurring at only about 2--3\% of generated tokens. To examine whether these flips actually correspond to harmful revisions, we replace the final-loop winner at each Late-Flip position with the most frequent alternative top-1 token over iterations 17--32, excluding the final-loop winner, and resume greedy decoding. 

As shown in \Cref{{tab:late_flip_reversion}}, this replacement can repair late-loop errors, but more often disrupts an otherwise correct trajectory. Thus, late-loop erroneous revisions do occur, but beneficial late revisions are more common, supporting the use of the final loop as the expert anchor on average.

\begin{table}[t]
\centering
\small
\begin{tabular}{lc}
\toprule
Outcome & Rate \\
\midrule
Incorrect $\rightarrow$ Correct & 9.9\% \\
Correct $\rightarrow$ Incorrect & 23.9\% \\
\bottomrule
\end{tabular}
\caption{Effect of replacing Late-Flip positions with alternative top-1 predictions.}
\vspace{-4mm}
\label{tab:late_flip_reversion}
\end{table}

\begin{table*}[t]
\centering
\setlength{\tabcolsep}{4.5pt}
\renewcommand{\arraystretch}{0.9}
\small
\begin{tabular*}{\textwidth}{@{\extracolsep{\fill}}llccccccccc}
\toprule
\multirow{2}{*}{\textbf{Model}} 
& \multirow{2}{*}{\textbf{Method}}
& \multicolumn{2}{c}{\textbf{GSM8K}}
& \multicolumn{2}{c}{\textbf{MATH-500}}
& \multicolumn{2}{c}{\textbf{HumanEval}}
& \multicolumn{2}{c}{\textbf{MBPP}}
& \multirow{2}{*}{\textbf{StrQA}}
\\
\cmidrule(lr){3-4}
\cmidrule(lr){5-6}
\cmidrule(lr){7-8}
\cmidrule(lr){9-10}
& 
& \textbf{Flex} & \textbf{Strict}
& \textbf{Flex} & \textbf{Strict}
& \textbf{HE} & \textbf{HE+}
& \textbf{MBPP} & \textbf{MBPP+}
& \\
\midrule
\multirow{3}{*}{\textbf{Huginn-0125}}
 & Greedy
        & 33.21
        & 23.12
        & 13.60 
        & 12.20
        & 24.39 
        & 20.73
        & 40.74 
        & 33.60 
        & 54.02

        \\

\cmidrule(lr){2-11}
& Ours  
        & \textbf{36.39} 
        & \textbf{24.34}
        & \textbf{15.60} 
        & \textbf{14.20}
        & \textbf{30.49} 
        & \textbf{27.44} 
        & 42.06
        & \textbf{36.77} 
        & \textbf{55.24}

\\
&    
        & \textcolor{green!50!black}{\textbf{+3.18}}  
        & \textcolor{green!50!black}{\textbf{+1.22}}
        & \textcolor{green!50!black}{\textbf{+2.00}}  
        & \textcolor{green!50!black}{\textbf{+2.00}}
        & \textcolor{green!50!black}{\textbf{+6.10}} 
        & \textcolor{green!50!black}{\textbf{+6.71}}  
        & \textcolor{green!50!black}{\textbf{+1.32}}  
        & \textcolor{green!50!black}{\textbf{+3.17}}  
        & \textcolor{green!50!black}{\textbf{+1.22}}  
\\

\midrule
\multirow{3}{*}{\textbf{Ouro-1.4B}}
 & Greedy
        & 78.70
        & 60.65
        & 50.20
        & 34.60
        & 69.50
        & 65.85
        & 72.75
        & 61.90
        & 64.00
        \\

\cmidrule(lr){2-11}
& Ours  
        & \textbf{79.83} 
        & \textbf{62.40}
        & 51.20 
        & \textbf{37.20}
        & \textbf{71.34} 
        & \textbf{67.68} 
        & \textbf{72.75} 
        & \textbf{62.17} 
         & \textbf{64.67}
\\
&    
        & \textcolor{green!50!black}{\textbf{+1.13}}  
        & \textcolor{green!50!black}{\textbf{+1.75}}  
        & \textcolor{green!50!black}{\textbf{+1.00}}  
        & \textcolor{green!50!black}{\textbf{+2.60}}  
        & \textcolor{green!50!black}{\textbf{+1.84}} 
        & \textcolor{green!50!black}{\textbf{+1.83}}  
        & \textcolor{green!50!black}{\textbf{+0.00}}  
        & \textcolor{green!50!black}{\textbf{+0.27}}  
        & \textcolor{green!50!black}{\textbf{+0.67}}  
\\
\bottomrule
\end{tabular*} \vspace{-2mm}
\caption{Performance comparison for Huginn and Ouro with fixed hyperparameters.}
\vspace{-2mm}
\label{tab:fixed_hyperparameters}
\end{table*}

\begin{table*}[t]
\centering
\setlength{\tabcolsep}{4.5pt}
\renewcommand{\arraystretch}{0.9}
\small
\begin{tabular*}{\textwidth}{@{\extracolsep{\fill}}lccccccc}
\toprule
\textbf{Method}
& \textbf{GSM8K}
& \textbf{MATH-500}
& \textbf{HE}
& \textbf{HE+}
& \textbf{MBPP}
& \textbf{MBPP+}
& \textbf{StrQA}
\\
\midrule

Greedy
    & 32.98 $\pm$ 0.97
    & 13.75 $\pm$ 0.71
    & 23.32 $\pm$ 1.46
    & 19.66 $\pm$ 1.46
    & 40.67 $\pm$ 0.63
    & 33.33 $\pm$ 0.60
    & 53.77 $\pm$ 0.07
\\

\midrule

LoopCD
    & \textbf{35.88 $\pm$ 0.72}
    & \textbf{16.00 $\pm$ 1.58}
    & \textbf{30.49 $\pm$ 1.37}
    & \textbf{27.59 $\pm$ 1.46}
    & \textbf{42.79 $\pm$ 1.05}
    & \textbf{35.32 $\pm$ 1.00}
    & \textbf{55.71 $\pm$ 0.68}
\\

& \textcolor{green!50!black}{\textbf{+2.90}}\phantom{ $\pm$ 0.72}
& \textcolor{green!50!black}{\textbf{+2.25}}\phantom{ $\pm$ 1.58}
& \textcolor{green!50!black}{\textbf{+7.17}}\phantom{ $\pm$ 1.37}
& \textcolor{green!50!black}{\textbf{+7.93}}\phantom{ $\pm$ 1.46}
& \textcolor{green!50!black}{\textbf{+2.12}}\phantom{ $\pm$ 1.05}
& \textcolor{green!50!black}{\textbf{+1.99}}\phantom{ $\pm$ 1.00}
& \textcolor{green!50!black}{\textbf{+1.94}}\phantom{ $\pm$ 0.68}
\\

\bottomrule
\end{tabular*}
\vspace{-2mm}
\caption{Performance of Huginn-0125 with LoopCD averaged over four seeds. Results are mean accuracy (\%) $\pm$ 95\% confidence intervals. GSM8K and MATH-500 use flex accuracy.}
\vspace{-4mm}
\label{tab:multi_seed}
\end{table*}

\subsection{Representative Failure Cases}

We next examine failures introduced by LoopCD itself, where Greedy
decoding would otherwise produce a correct trajectory.

\paragraph{CASE 1: Unnecessary intervention on an easy token.}
Although LoopCD disproportionately affects hard tokens, its implicit
selectivity is not perfect. For example:

\noindent\textbf{Greedy [Correct]:}
$\ldots 43 \times 3 = 129 \ldots$\\
\textbf{LoopCD [Incorrect]:}
$\ldots 43 \times 3 = 1{,}293 \ldots$

At the divergent position, both the expert and amateur prefer the correct token \texttt{2}, but the amateur assigns it slightly higher confidence. The contrastive term therefore over-penalizes \texttt{2} and promotes the comma token \texttt{","}, thereby overturning an otherwise correct easy token prediction. Consistent with \Cref{appendix:finding_2}, such easy token margin flips are rare (0.82\%).

\paragraph{CASE 2: Initial-token trajectory divergence.}
LoopCD can also change the initial generated token and thereby redirect
the subsequent reasoning trajectory:

\noindent\textbf{Greedy [Correct]:} \\
"The original piece of wire was 4 feet long $\ldots$"\\
\textbf{LoopCD [Incorrect]:} \\
"4 feet is $4 \times 12 = 48$ inches $\ldots$"

Here, the initial token changes from \texttt{The} to \texttt{4}. Because subsequent predictions are conditioned on the generated prefix, this early intervention leads to a different reasoning path and ultimately an incorrect answer. Such initial-token divergence occurs in approximately 1.2\% of the evaluated trajectories.

Overall, beneficial corrections remain more common: on GSM8K with Huginn, the ratio of incorrect-to-correct changes to correct-to-incorrect changes is 1.49.

\section{Robustness of LoopCD}
\label{sec:robustness_loopcd}
\subsection{Fixed-Hyperparameter Results}

We evaluate whether LoopCD remains effective without benchmark-specific
tuning. For each model, we use one fixed configuration across all
benchmarks: $\ell_a=8$ and $\lambda=0.3$ for Huginn, and $\ell_a =1$ and
$\lambda=0.2$ for Ouro.

As shown in \Cref{tab:fixed_hyperparameters}, LoopCD improves over Greedy
decoding on all benchmarks except MBPP with Ouro, where it matches Greedy.
This suggests that LoopCD is robust to fixed, benchmark-agnostic
hyperparameter settings.

\subsection{Multi-Seed Evaluation}
\label{appendix:multi_seed}
To further assess the robustness of LoopCD, we evaluate Huginn across four random seeds. Huginn initializes its recurrent state with Gaussian noise, so different seeds induce different latent trajectories even under greedy decoding. 
\Cref{tab:multi_seed} reports the mean accuracy and 95\% confidence intervals across the four seeds. LoopCD improves the mean performance over Greedy across all benchmarks. These results show that the improvements of LoopCD are robust to variations in the randomized initial recurrent state.

\section{Qualitative Examples}
\label{sec:qualitative}
\begin{figure*}[p]
    \centering

    \begin{subfigure}{\textwidth}
        \centering
        \includegraphics[width=\textwidth,height=0.42\textheight,keepaspectratio]{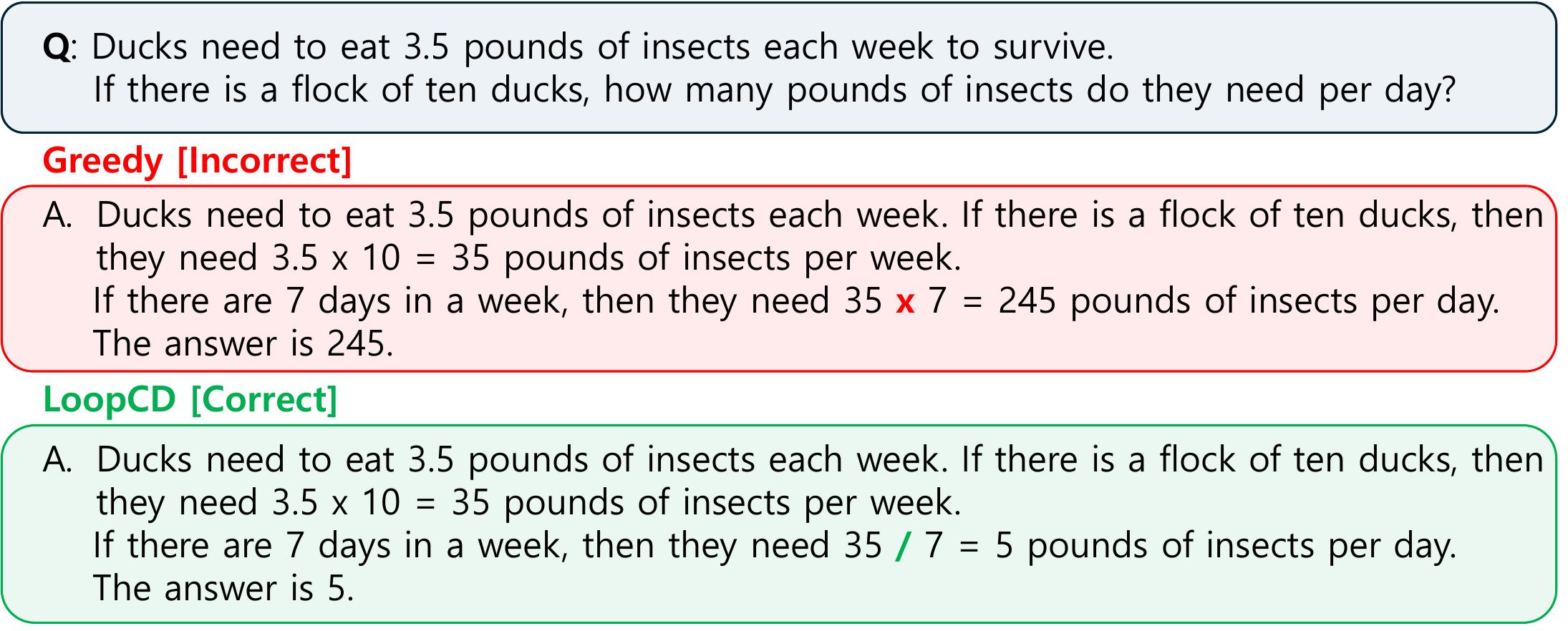}
    \end{subfigure}

    \vspace{15mm}

    \begin{subfigure}{\textwidth}
        \centering
        \includegraphics[width=\textwidth,height=0.42\textheight,keepaspectratio]{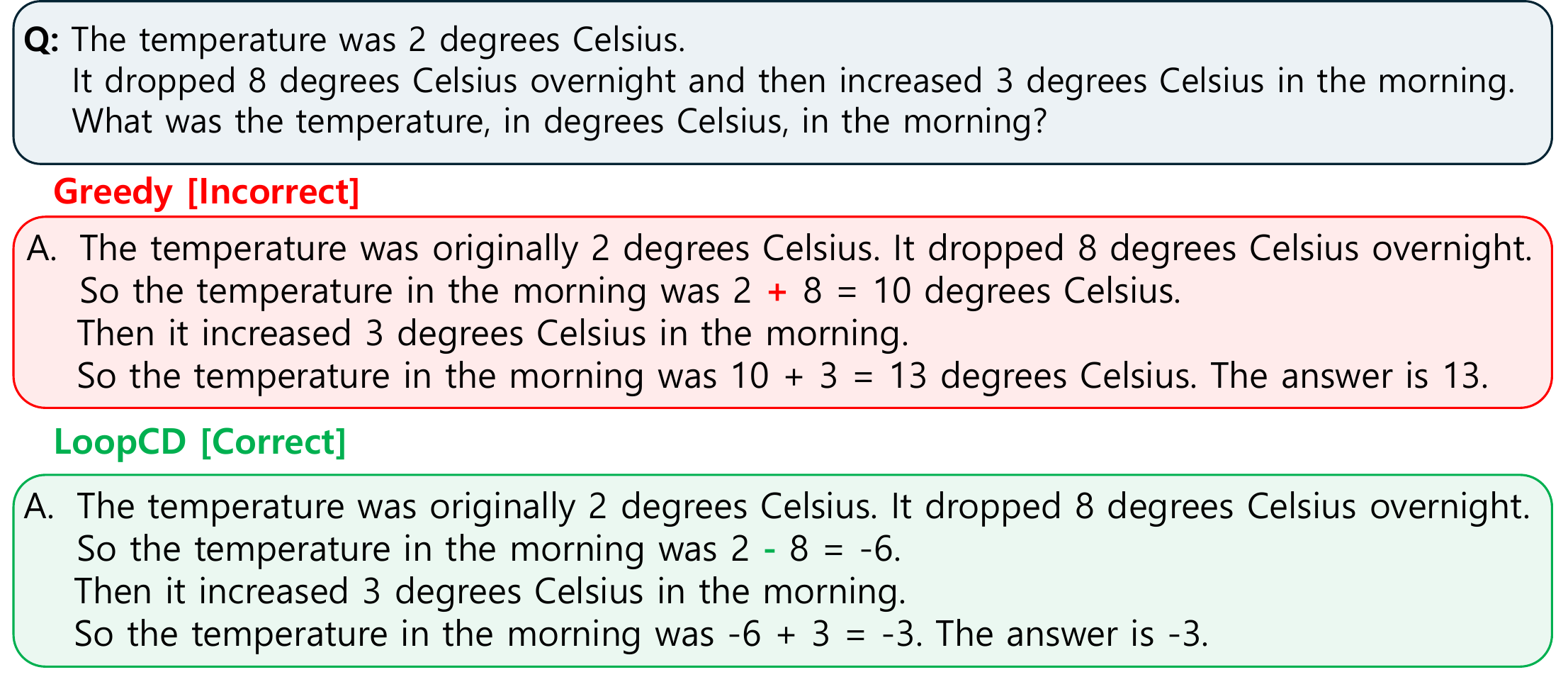}
    \end{subfigure}

    \caption{
    Qualitative comparison between Greedy decoding and LoopCD.
    Tokens highlighted in red and green indicate positions where the generations of Greedy decoding and LoopCD diverge.
    LoopCD adjusts these reasoning-critical tokens, leading to more consistent reasoning trajectories and correct final answers.
    }
    \label{fig:quality}
\end{figure*}

We present qualitative examples comparing Greedy decoding and LoopCD in \Cref{fig:quality}.
The highlighted tokens indicate positions where the two decoding methods diverge.
These examples show that LoopCD often modifies reasoning-critical tokens, leading to improved reasoning outcomes.

\end{document}